%% file: acl2025.tex
\documentclass[11pt]{article}
\usepackage[final]{acl}

\usepackage{polyglossia}

\setmainlanguage{english}
\setotherlanguage{hebrew}

\newfontfamily\hebrewfont[Script=Hebrew,Scale=MatchLowercase]{FreeSerif.ttf}

\usepackage{microtype}      
\usepackage{graphicx} 
\usepackage{booktabs}   
\usepackage{enumitem}       
\usepackage{multirow}       
\usepackage{adjustbox}      
\usepackage{algorithm}
\usepackage{algorithmic}
\usepackage{placeins}
\usepackage{float}

\usepackage[
    skins,
    breakable,
    xparse,
    fitting,
    listings,
    theorems,
    documentation
]{tcolorbox}

\title{Beyond Word Boundaries: A Hebrew Coreference Benchmark and an Evaluation Protocol for  Morphologically Complex Text
        }

\author{
Refael Shaked Greenfeld \\
Bar-Ilan University \\
\texttt{shakedgreenfeld@gmail.com}
\And
Reut Tsarfaty \\
Bar-Ilan University \\
\texttt{reut.tsarfaty@biu.ac.il}
}

\newfontfamily\hebrewfonttt[Script=Hebrew,Scale=MatchLowercase]{MiriamMonoCLM-Book.ttf}
\begin{document}
\maketitle

\begin{abstract}

Coreference Resolution (CR) is a fundamental NLP task critical for long-form tasks as information extraction, summarization, and many business applications. However, CR methods originally designed for English struggle with Morphologically Rich Languages (MRLs), where mention boundaries  do not necessarily align with word boundaries, and a single token may consist of multiple anaphors. CR modeling and evaluation protocols standardly assume that, as in English, words and mentions mostly align. However, this assumption breaks down in MRLs, 
particularly  in the context of LLMs' raw-text processing and end-to-end tasks.  To  assess and address this challenge, we introduce {\em KibutzR}, the first comprehensive CR dataset for Modern Hebrew, an MRL rich with complex words and pronominal clitics. We deliver an annotated dataset that identifies mentions at  word, sub-word and multi-word levels, and propose  an evaluation protocol that directly addresses word/morpheme boundary discrepancies. Our experiments show that contemporary LLMs perform significantly worse on Hebrew than on English, and that performance degrades  on raw unsegmented text. Crucially, we show an inverse performance-trend in Hebrew relative to English, where smaller encoders perform far better than contemporary decoder models, leaving ample space for investigation and  improvement. We deliver a new benchmark for Hebrew coreference resolution and a segmentation-aware evaluation protocol to inform future work on other MRLs.

 \end{abstract}

\section{Introduction}

The field of Natural Language Processing has achieved remarkable breakthroughs in recent years, driven by advanced large language models (LLMs) and large-scale resources. 

Despite these advances, complex word structures continue to pose significant challenges for automatic text processing and discourse-level understanding.
In particular, the task of Coreference Resolution (CR) involves identifying and clustering entities across a discourse, enabling deeper text comprehension. However, achieving accurate CR   presents unique challenges in  {\em morphologically rich languages} (MRLs), with phenomena as pronominal clitics and morpheme-stacking that obscure mention boundaries. These linguistic intricacies make mention detection and coreference resolution particularly difficult in MRLs.

Modern Hebrew, a Semitic morphologically rich language,  exemplifies these challenges,  where a single token in Hebrew may consist of multiple   anaphors that designate different entities in one and the same token, via construct-state nouns and pronominal clitics \cite{more2016data}. 
 Similarly, Arabic dialects show  pronominal clitic fusion \cite{maamouri2004penn}, Turkish stacks multiple morphemes creating boundary detection problems \cite{schuller2017marmara}, and Slavic languages encode multiple grammatical relations within a single word. All these languages share the same critical challenge --- that standard CR models and metrics assume that  mentions  essentially align with word boundaries --- but this alignment either doesn't  naturally  apply, or requires error-prone preprocessing.

 As NLP shifts towards end-to-end architectures and LLM processing of raw texts, evaluation practices designed for CR in English turn out to fail on MRL texts.
Concretely, the shift to generative LLMs has created the following evaluation gap: these models process raw texts directly, but we lack frameworks to assess performance when morphological segmentation and coreference errors are intertwined. 

Exacerbating this, 
many lower-resourced MRLs still lack even the very basic research infrastructure for CR, including public benchmarks to assess CR and track empirical progress.
Without benchmarks that isolate the error sources, we cannot diagnose where and why CR models fail on morphologically-rich texts and suggest mitigation.

This work focuses on Hebrew, an MRL rich with  pronominal clitics that break word-to-mention alignment.
While significant progress has been made in Hebrew via pre-trained  encoders such as AlephBERT \cite{seker2022alephbert} and DictaBERT \cite{shmidman2023dictabert}, as well as Large Language Models (LLMs) like DictaLM \cite{shmidman2024adaptingllmshebrewunveiling}, and
%
while the Hebrew NLP community contributed valuable resources for shorter-text tasks such as sentiment analysis, named entity recognition (NER) \citep{Bareket_2021_nemo}, and  question answering (QA) \citep{cohen2025heqlargediversehebrew}, there remains a notable lack of resources for {\em discourse-level} understanding, and in particular {\em coreference resolution} (CR), limiting the development of more complex applications and long-form Hebrew tasks.


In this work we address the multifaceted CR challenge in Hebrew 
by developing the first comprehensive Modern Hebrew CR dataset, \textit{KibutzR}, accompanied by annotation guidelines that account for morphologically complex phenomena and discrepancies in word-mention boundaries. 
Additionally we introduce an evaluation protocol that remains sensitive to word-mention boundary discrepancies, 
providing a robust framework for comparing and contrasting  models, both
generative decoders and encoder-based, 
on raw (unsegmented) texts.

The contribution of this paper is thus manifold. First, we present {\em KibutzR}, the first modern Hebrew coreference resolution dataset, alongside detailed annotation guidelines and a rule-based mention detector. Second, we introduce a segmentation-aware evaluation protocol that makes word--mention boundary discrepancies explicit without assuming gold morphological analysis. Although instantiated for Hebrew, this protocol provides a clear blueprint for porting segmentation-aware coreference evaluation to other MRLs. Finally, we show a comprehensive empirical analysis of contemporary models tracing performance gaps back to their roots in detection and clustering. Together, these contributions provide an immediate resource for Hebrew NLP, and a methodological blueprint for targeting and achieving improved CR capabilities in Hebrew and other MRLs.

\section{Morphological Challenges in  Coreference Resolution}
\label{sec:mrl-morphology}

\subsection{Referring Expressions  in MRLs}
In morphologically rich languages, referring expressions frequently occur as subtoken units. In Hebrew,  for instance, the token  \texthebrew{דבריו} (`his words') exemplifies this challenge: this single orthographic token requires segmentation into \texthebrew{דבר\_של\_הוא} (words\_of\_he).\footnote{By Hebrew UD v2 conventions \citep{sade-etal-2018-hebrew}.} It thus contains two separate mentions that can co refer to distinct entities \textit{words} and \textit{he}. 

This phenomenon extends across typologically diverse languages. Arabic \textit{kitābu-hu} ('his book') similarly fuses nominal and pronominal elements that must be decomposed into \textit{kitāb} + \textit{hu} for coreference resolution \citep{maamouri2004penn}. Turkish agglutination produces \textit{evlerimizden} (`from our houses'), stacking morphemes as   \textit{ev}+\textit{ler}+\textit{imiz}+\textit{den}, where possessive and plural markers create overlapping mention spans \citep{schuller2017marmara}. Basque \textit{etxekoak} (`those of the house') demonstrates comparable fusion through \textit{etxe}+\textit{ko}+\textit{ak}, interleaving genitive marking with determination \citep{soraluze2019euskor}. Georgian verbal morphology presents the most complex case: \textit{mogvts'eren} (`they will write to us') encodes multiple argument references through preverb \textit{mo-}, object marker \textit{gv}, and subject agreement \textit{-en}, all within a single form.
These morphological patterns fundamentally challenge the word-as-unit assumption underlying most CR systems. Unlike English, where mention boundaries align with whitespaces, MRLs require models to simultaneously segment morphemes and resolve their referential relations, transforming a primarily semantic task into one demanding morphosyntactic proficiency.

Beyond such bounded clitics, three additional phenomena increase the difficulty of CR in  MRLs. First, pro-drop significantly increases ambiguity by omitting arguments that can be recovered from morphological cues \citep{demir2024mention,maamouri2009conventional,soraluze2019euskor}. Unlike languages with obligatory overt subjects, pro-drop languages force CR systems to infer  referents from verbal inflections, creating additional decision points in the resolution process. 

Second, {\em construct-state nouns} (CSNs)
create deeply nested nominal structures that challenge mention boundary detection. Unlike English, which uses prepositions and determiners to mark possession and modification, CSN concatenates nouns directly, creating complex multi-word expressions as \texthebrew{דו"ח ישיבת ועדת מנויי דיני בית הדין הרבני הגדול} (`report of the meeting of the committee of appointments of judges of the Great Rabbinical Court'). These constructions pose significant challenges for CR: they need to identify the entire construct chain as a single mention while also recognizing potential embedded mentions to (co)-refer to.

Third,  many MRLs show flexible word order, that erodes positional heuristics for salience and proximity that systems traditionally exploit \citep{althubaity2017coreference,soraluze2019euskor}.
Classical approaches relied on surface positioning patterns from fixed word order languages, where syntactic roles correlate with linear position. However, the equivalence of Hebrew \texthebrew{דן קרא את הספר} and \texthebrew{את הספר קרא דן} (`Dan read the book'), undermines  positional features and   surface-level cues.
%

These three factors strip away surface cues that English-centric systems exploit, forcing (any kind of) models to rely more on morphological  and semantic understanding rather than mere surface patterns. 

\subsection{Word Segmentation: From Pipeline Artifacts to Modern Bottlenecks}
\label{ssec:segmentation-challenge}

A long-standing, implicit yet persistent assumption in evaluating CR systems is that space-delimited tokens represent single mentions.
This  paradigm, while standard in the field, fundamentally misrepresents the challenge of MRL CR and creates a substantial gap between reported performance in research papers and real-world deployment scenarios.

In English, where  
referential expressions often appear as standalone tokens, the distinction is inconsequential --- models trained on space-delimited text face no additional challenge in deployment. 
In MRLs,  the  challenge to identify and segment referential material is inseparable from coreference resolution, as multiple referential material routinely appear within a complex word that are input-streamed to the model.

To address this intrinsic duality  of mention detection and coreference clustering, MRL research standardized 
the use of gold-segmented text for coreference annotation and coreference evaluation as a pragmatic solution. 
Concretely, all major MRL CR corpora adopt this approach: the OntoNotes–Arabic corpus reuses the Penn ATB segmentation; Marmara–Turkish inherits METU–Sabancı morphological analysis; and EUSKOREC pre-extracts mentions with finite-state rules \citep{pradhan-etal-2012-conll,schuller2017marmara,soraluze2019euskor}. While gold segmentation simplifies annotation and streamlines evaluation, it creates an artificial evaluation scenario that diverges from real-world text  processing.

In the earlier NLP-pipelines era,  {\em mention detection} and {coreference clustering} formed two distinct phases, allowing researchers to isolate error sources by comparing performance with gold versus automatic mention-boundary detection. This diagnostic capability has been crucial: Marmara's baseline dropped by 31.4 CoNLL F\textsubscript{1} when switching from gold to automatic mentions, while Basque reported 19–21-point drops \citep{schuller2017marmara,soraluze2015adapting}. Such comparisons revealed precisely where systems failed: was the bottleneck in  the {\em detection} or {\em clustering} phase?

In today's neural end-to-end era, LLMs process raw text directly, fusing mention detection and coreference clustering into a single phase \citep{lee-etal-2017-end,joshi-etal-2019-bert}. The relevant contrast is 
in what the model should be provided as input: should this be raw text or pre-segmented tokens? Subsequently, the question becomes: when a CR model fails on raw MRL texts,  can we distinguish whether errors stem from segmentation challenges (failing to identify  clitics), linguistic nuances (mishandling construct states), or genuine coreference confusion (incorrect cluster)?

This shift fundamentally changes how we should evaluate MRL coreference systems, yet existing CR benchmarks for MRLs continue to report results on gold-segmented text, creating an evaluation-deployment gap: models achieve strong performance on pre-segmented benchmarks but struggle with raw text in deployment. This misalignment prevents us from understanding --- let alone improving --- actual CR performance on MRLs, as we lack diagnostic tools to isolate  error sources and quantify their relative impact.

\subsection{Towards Resolving Coreference Resolution in MRLs}
\label{ssec:gaps}

To address the fundamental gap in MRL CR  we propose a comprehensive solution that aligns research benchmarks with deployment realities while maintaining diagnostic capabilities, via 
two interconnected objectives.

First, we need a comprehensive dataset that captures the full spectrum of mention spans, 
from subword to multiword levels, to reflect how referring expressions actually manifest in these languages. Unlike existing MRL corpora that rely on pre-existing segmentation schemas, we need annotations that explicitly mark referential material within complex word-forms --- pronominal clitics, construct states, and other linguistic phenomena. We address this gap in Section~\ref{sec:corpus-construction} by constructing KibutzR, the first comprehensive Hebrew coreference dataset with morpheme-aware annotations that systematically handle  morphological challenges in CR.

Second, we critically need evaluation scenarios that bridge the gap between research and deployment while preserving diagnostic capabilities. Rather than abandoning the insights from pipeline-era evaluations, we propose a diagnostic ladder of input conditions that systematically varies the level of  preprocessing: (1) raw text, as encountered in real applications; (2) automatic segmentation, revealing the impact of segmentation errors; and (3) gold segmentation, isolating pure coreference challenges from  morphological intricacies. This three-regime evaluation protocol allows us to quantify the relative contribution of different error sources --- segmentation versus clustering --- while maintaining comparability with existing benchmarks.
%
%
By testing models across all three input conditions, we can answer critical questions about model capabilities: How substantial is the performance gap between raw and segmented text? What proportion of errors stem from segmentation versus genuine coreference confusion? Do modeling advantages observed in English, where LLMs now dominate, translate to MRLs when processing raw text?
We define and implement this evaluation framework in Section~\ref{sec:experiments}.


With these foundations in place --- comprehensive data and diagnostic evaluation --- in Section~\ref{sec:resuls} we establish baseline performance across architectures, both state-of-the-art generative LLMs and neural encoder models, providing the first comprehensive evaluation leaderboard for Hebrew coreference resolution that reflects the gap between research scenarios and deployment realities.

\section{Building KibutzR: Construction and Annotation of the Modern Hebrew Coreference Corpus}
\label{sec:corpus-construction}

\subsection{Scope and Document Selection}

\begin{figure}[t]
    \centering
    \includegraphics[width=.8\columnwidth]{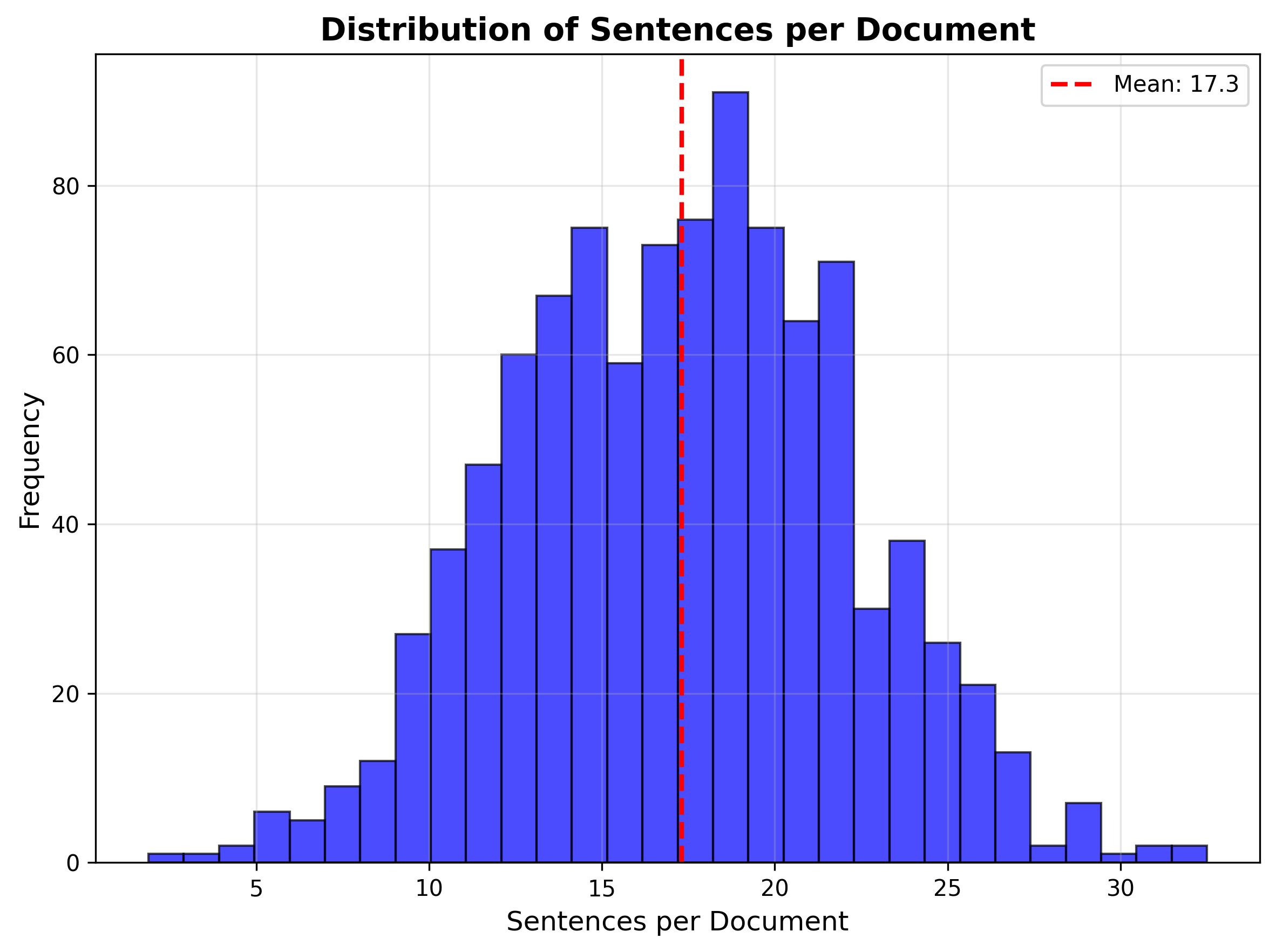}
    \caption{Distribution of sentences per document.}
    \label{fig:sent-dist}
\end{figure}

The Hebrew KibutzR dataset is based on The Hebrew Universal Dependencies Treebank \citep{sade-etal-2018-hebrew}, containing 6,151 sentences,  without document boundaries --- a critical limitation for coreference resolution research.
To solve this we reconstructed the original document structure. Through metadata analysis and discourse pattern recognition, we successfully identified document boundaries and segmented the continuous sentence stream into 351 complete documents. This transformation --- from isolated sentences to coherent documents averaging 17.3 sentences (453.6 tokens)  As shown in Figure~\ref{fig:sent-dist} --- provides Hebrew NLP with its first document-aware corpus derived from the richly annotated UD treebank.

The reconstruction process required rethinking train/dev/test partitioning. The original Hebrew UD splits scattered sentences from the same document across different partitions, creating evaluation contamination. We therefore relocated any document appearing in multiple partitions exclusively to training. This principled partitioning preserves compatibility with existing Hebrew NLP tools trained on UD while ensuring clean evaluation: dependency parsers and morphological analyzers can operate on the same distribution without compromising coreference evaluation integrity.


\begin{table}[t]
\centering
\small
\scalebox{0.9}{\begin{tabular}{lrrrrr}
\toprule
\textbf{Split} & \textbf{Docs} & \textbf{Sents} & \textbf{Tokens} & \textbf{Mentions} & \textbf{\%Docs}\\
\midrule
Train & 301 & 5,236 & 137,333 & 17,500 & 85.8 \\
Dev   &  26 &   428 &  10,474 &  1,243 &  7.4 \\
Test  &  24 &   487 &  12,168 &  1,451 &  6.8 \\
\midrule
Total & 351 & 6,151 & 159,975 & 20,194$^\dagger$ & 100 \\
\bottomrule
\end{tabular}}
\caption{Corpus statistics for KibutzR. The marking  $^\dagger$ excludes singletons. With singletons, the corpus contains 47,879 mentions.}

\label{tab:corpus-stats}
\end{table}


\subsection{Annotation Guidelines}
Our annotation guidelines adopt OntoNotes 5.0 as their foundation while making two substantive departures to accommodate concrete morphosyntactic phenomena (Sec. \ref{sec:mrl-morphology}). 

First, OntoNotes restricts mention boundaries to space-delimited tokens and does not annotate sub-token morphemes. 
This is problematic for Hebrew where possessive suffixes and pronominal clitics carry independent referents. We therefore permit \emph{morpheme-level} mentions within orthographic tokens whenever fused morphemes carry reference. To illustrate, in \texthebrew{דבריו} ('his words'), we annotate the possessive suffix as a separate pronominal mention alongside the full token. This modification enables annotation of possessive and pronominal clitics, and proclitic prepositions/conjunctions --- all central to Hebrew reference. We do not  split lexical roots from templatic patterns.

Second, OntoNotes generally annotates only single maximal NPs (no nested mentions/i-within-i), with limited exceptions (e.g., proper-noun premodifiers and appositives) --- insufficient for  recursive construct state nouns  where sub-constituents maintain independent referents. We therefore treat recursive \emph{smixut} (construct state) as nested mention hierarchies. For \texthebrew{ממשלת אנגליה} (`the government of England'), we allow coreference links to both the full compound and the embedded constituent \texthebrew{אנגליה} (`England'), capturing the dual referential nature of these constructions.\footnote{Other adjustments, e.g.\ head selection for quantificational/partitive NPs and Hebrew-specific tests for non-referentials, are refinements rather than conceptual.}

\begin{figure}[t]
    \centering
    \includegraphics[width=.8\columnwidth]{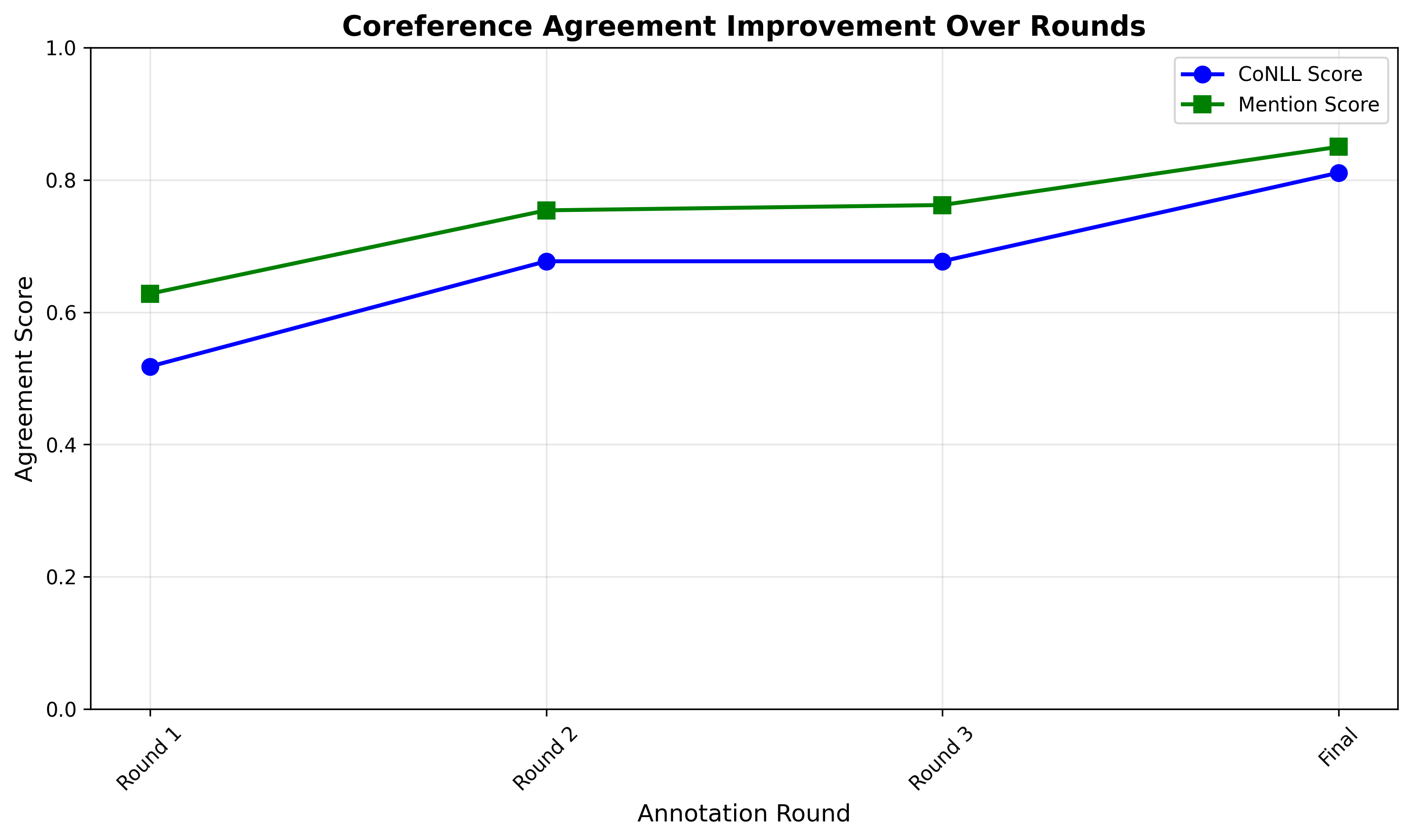}    
    \caption{Agreement improvement across annotation rounds (CoNLL and Mention F$_1$).}
    \label{fig:pairwise}
\end{figure}

The complete annotation manual, with detailed examples, is available in our repository.\footnote{Data, guidelines, and metadata are available at \url{https://github.com/OnlpLab/hebrew_coreference_data/}.}

\subsection{Annotation Pipeline and Quality Control}
\begin{figure}[t]
    \centering
    \includegraphics[width=.95\columnwidth]{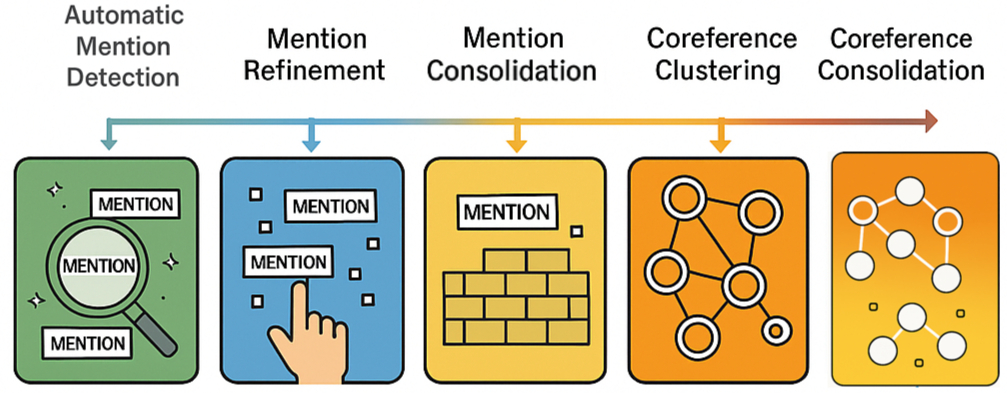}
    \caption{The five-stage annotation pipeline}
    \label{fig:annotation-pipeline}
\end{figure}


Our annotation methodology follows a systematic five-stage pipeline, illustrated in Figure~\ref{fig:annotation-pipeline}.
%
Given the challenges outlined in Section~\ref{sec:mrl-morphology}, 
 we separate mention boundary decisions from coreference clustering. This separation allows annotators to first resolve complex morphological boundaries 
 before making referential judgments, reducing cognitive load and improving consistency.
%
All stages are carried out independently by at least three annotators to secure reliable coverage in high agreement. 

For the annotation platform, we extended the TNE annotation platform \citep{elazar-etal-2022-text} with support for morpheme-level span editing and robust right-to-left Hebrew display, enabling efficient and accurate annotation of complex Hebrew morphological structures. Annotators were compensated at 50 NIS/hour (60\% more than the state's minimum wage).

The process begins with \textbf{automatic mention detection}.  We designed a custom rule-based mention detector that leverages Universal Dependencies parses with 
rules for clitics, construct states, and other morphological and morphosyntactic phenomena. This recall-oriented system processes raw Hebrew text from UD-annotated documents and produces text with candidate mention-spans pre-marked, including nested mentions. The high-recall design ensures comprehensive candidate coverage, allowing annotators to focus on refinement rather than mention discovery.

Next is the \textbf{mention refinement stage}, where human annotators receive text with pre-marked mention candidates and systematically accept or reject each candidate while adding any mentions missed by the rule-based detector. Using morpheme-level span editing capabilities, annotators can precisely define mention boundaries that cross traditional word boundaries. 
Each annotator produces their individual mention decisions for subsequent consolidation.

Later, the \textbf{mention consolidation stage} resolves inter-annotator variation by merging mention decisions into a unified inventory. Boundary disagreements are settled by majority vote or expert adjudication, producing a fixed set of mentions before clustering.

During \textbf{coreference clustering}, annotators process the consolidated mentions sequentially using a single-link strategy, deciding for each mention whether it opens a new cluster or links to an existing one. Singletons are retained to ensure complete mention coverage.

Finally, the {\bf quality control and adjudication stage} handles systematic disagreement resolution through expert consolidation. 
An expert annotator reviews all disagreements and consolidates them cluster by cluster in sequential order, following an approach similar to   \citet{bornstein-etal-2020-corefi}. This expert makes final decisions on conflicting annotations based on linguistic criteria and annotation guidelines, ensuring consistency across the dataset. 

We monitored {\bf inter-annotator agreement} after each batch and conducted a targeted revision following the first three batches. This iterative process improved macro average pairwise scores from 0.63 (mention) / 0.52 (CoNLL) initially to a final agreement of 0.87 for mentions and 0.81 for CoNLL F$_1$ --- relative gains of 38\% and 56\% respectively (Figure~\ref{fig:pairwise}). The final agreement scores for the dataset are: Mention F$_1$ = 0.87 (P: 0.85, R: 0.90), MUC = 0.84, B$^3$ = 0.81, CEAF$_\phi$ = 0.79. 

Notably, these final agreement scores {\bf match or surpass} all of those reported for the most known large-scale CR corpora: OntoNotes (85.8\% MUC F$_1$; \citealt{pradhan-etal-2012-conll}), LitBank (78\% B$^3$ F$_1$; \citealt{bamman-etal-2019-large}), and PreCo (77\% agreement; \citealt{chen-etal-2018-preco}).

\section{Experimental Setup}
\label{sec:experiments}



\paragraph{Goal and Evaluation Scenarios} We set out to evaluate coreference resolution models and isolate the impact of  morphological complexity on downstream performance. To do so, we propose to assess CR  in 3 input conditions:
\begin{itemize}[leftmargin=1.5em]
\item \textbf{Raw Text:} The CR models receive unsegmented text and must discover mention boundaries and resolve coreference chains
\item \textbf{Automatic Segmentation:} The CR models operate on text segmented by a state-of-the-art linguistic segmentation model.\footnote{For the automatic segmentation condition, we use the joint Hebrew segmentation model of \citet{yshaayahu-levi-tsarfaty-2024-truly}, which achieves state-of-the-art performance on Hebrew UD benchmarks and produces Universal Dependencies-conformant boundaries -- 98.52 F1 on Hebrew UD benchmarks \cite{yshaayahu-levi-tsarfaty-2024-truly}.}
\item \textbf{Gold Segmentation:} The CR models receive gold-standard segments boundaries.
\end{itemize}
For LLMs, we additionally assess CR performance in a \textbf{Gold Mentions} condition, where models receive pre-identified mentions, to isolate  performance on the {\em clustering} subtask.\footnote{
Because neural encoders operate on token sequences \emph{(i.e., as token/span classifiers rather than segmenters)}, they require pre-segmented input by  design and cannot process raw text directly,
so neural encoder based systems are evaluated in automatic and gold segmentation conditions only. 
while LLMs are evaluated across all three input conditions plus gold mentions. This asymmetry reflects fundamental architectural differences: current neural encoder based systems inherit pipeline assumptions where segmentation precedes coreference, while LLMs are capable of handling both tasks simultaneously from raw text.}

\paragraph{Metrics.} 
We follow the standard CoNLL-2012 shared task evaluation setup \citep{pradhan-etal-2012-conll}, which established  three metrics as the canonical framework for cross-linguistic coreference evaluation.\footnote{Although we annotated singleton mentions, all reported results exclude them, following the conventions of prior work \citep{cattan-etal-2021-realistic}.}
We report standard CR metrics: MUC \citep{vilain-etal-1995-model}, B$^3$ \citep{bagga-baldwin-1998-entity-based}, and CEAF$_\phi$ \citep{luo-2005-coreference}, along with their arithmetic mean (CoNLL F$_1$). MUC measures link-based precision and recall, B$^3$ evaluates mention-based clustering quality, and CEAF$_\phi$ computes the optimal alignment between predicted and gold clusters.

\paragraph{Neural Encoder Models.} We fine-tune two state-of-the-art neural CR architectures on KibutzR's training set. The \emph{wl-coref} system \citep{dobrovolskii-2021-word} uses word-level span representations with a coarse-to-fine antecedent scoring mechanism. The \emph{LingMess} system \citep{otmazgin-etal-2023-lingmess} incorporates multiple expert scorers that capture different linguistic signals. We adapt LingMess to Hebrew (LingMess-He) by replacing the English stopword and pronoun lists with their Hebrew equivalents in the scorer modules. We experiment with two Hebrew neural encoders, AlephBERT \citep{seker2022alephbert} and DictaBERT \citep{shmidman2023dictabert}, which  constitute the strongest pretrained models available for Modern Hebrew to date.\footnote{Training: 150 epochs, AdamW optimizer (encoder lr=$1 \times 10^{-5}$, task lr=$3 \times 10^{-4}$), dropout=0.3. \texttt{wl-coref}: max span=64, top-k=50. LingMess-He: max span=30, top-$\lambda$=0.4. Results averaged over 5 seeds.All models use early stopping on development CoNLL F$_1$ with hyperparameters following the original papers. Full hyperparameter details are supplied in the Appendix \ref{app:neural-hyperparams}. All code and training scripts are available at \url{https://github.com/OnlpLab/hebrew_coreference}.}

\paragraph{Generative Language Models.} We evaluate eight state-of-the-art LLMs:  a Hebrew-monolingual model, DictaLM 2.0 \citep{shmidman2024adaptingllmshebrewunveiling}
, and major multilingual LLMs: GPT-4.1/4o/o1/o3, and Gemini 2.0-Flash/-Lite/2.5-Pro. 
Following prior work \citep{le2023largelanguagemodelsrobust}, we use zero-shot prompting for the gold mentions condition to enable direct comparison with their English setup. For the end-to-end  task (i.e., raw and gold segmentation conditions), we employ 2-shot prompting with examples from the training data, as zero-shot evaluation proved ineffective for this more complex task. We experimented with multiple prompting strategies to optimize model performance. Our final reported end-to-end prompt incorporates a Chain-of-Thought (CoT) structure through a three-step pipeline that explicitly handles Hebrew morphological challenges. We evaluated each scenario with temperature~0 and report averages across five runs. See  complete prompts and examples in appendix \ref{app:prompts}.




\section{Results and Discussion}
\label{sec:resuls}




\paragraph{Neural Encoder-Based Systems}

\begin{table}[t]
\centering
\begin{adjustbox}{max width=\columnwidth}
\begin{tabular}{lcccc}
\toprule
\multirow{2}{*}{\textbf{Model}} &
\multicolumn{4}{c}{\textbf{Gold Segmentation}} \\
\cmidrule(lr){2-5}
 & MUC & B$^{3}$ & CEAF$_\phi$ & CoNLL F$_1$ \\
\midrule
wl-coref (+AlephBERT)         & 47.1 & 41.6 & 44.4 & 44.4 \\
wl-coref (+DictaBERT-base)    & 47.7 & 41.9 & 44.1 & 44.5 \\
lingmess-he (+AlephBERT)      & 45.7 & 41.8 & 45.2 & 44.3 \\
lingmess-he (+DictaBERT)      & \textbf{52.6} & \textbf{47.7} & \textbf{51.0} & \textbf{50.4} \\
\bottomrule
\end{tabular}
\end{adjustbox}
\caption{neural encoder baseline performance with gold segmentation. Results averaged across 5 seeds; standard deviations range from $\pm$0.7 to $\pm$3.7 F$_1$ points. This evaluation regime matches all prior coreference work on other languages.}
\label{tab:neural}
\end{table}

Table~\ref{tab:neural} presents our neural encoder baseline results under gold segmentation, the standard de facto evaluation standard in coreference resolution. 
Our strongest system, \emph{LingMess-He} with the \emph{DictaBERT} encoder, achieves 50.4 CoNLL F$_1$. 
The choice of the Hebrew encoder also proves critical; DictaBERT consistently outperforms AlephBERT by 5--6 F$_1$ points, likely due to its larger pre-training corpus and larger vocabulary, optimized for Modern Hebrew.


\begin{table*}[t]
  \centering
  \begin{adjustbox}{max width=\textwidth}
    \begin{tabular}{lccccccccc}
      \toprule
      \multirow{2}{*}{\textbf{Model}} &
      \multicolumn{4}{c}{\textbf{Gold Segmentation}} &
      \multicolumn{4}{c}{\textbf{SOTA Automatic Segmentation}} &
      \multirow{2}{*}{\textbf{$\Delta$F$_1$}} \\
      \cmidrule(lr){2-5}\cmidrule(lr){6-9}
       & MUC & B$^{3}$ & CEAF$_\phi$ & F$_1$ &
         MUC & B$^{3}$ & CEAF$_\phi$ & F$_1$ & \\
      \midrule
      lingmess-he (+AlephBERT)    & 45.7 & 41.8 & 45.2 & 44.3 & 44.2 & 40.4 & 43.9 & 42.8 & $-$1.4 \\
      lingmess-he (+DictaBERT)    & 52.6 & 47.7 & 51.0 & 50.4 & 50.0 & 45.1 & 48.5 & 47.9 & $-$2.6 \\
      \bottomrule
    \end{tabular}
  \end{adjustbox}
  \caption{Neural model performance under gold versus automatic segmentation. The $\Delta$F$_1$ column shows absolute performance drop when using SOTA segmentation instead of gold tokens. Both conditions use identical text; only token boundaries differ.}
  \label{tab:segmentation-comparison}
\end{table*}

To isolate the effect of linguistic segmentation on downstream performance, we evaluated our best models on gold-segmented input vs.\ predicted segmentation.
Table~\ref{tab:segmentation-comparison} 
reveals a consistent performance drop of 1.4--2.6 F$_1$ points when replacing gold with automatic segmentation.  Although modest in absolute terms, this degradation 
isolates segmentation as an independent bottleneck in Hebrew NLP pipelines. This finding reveals that reporting scores only on gold-segmentation scenarios, hides a substantial portion of the error budget. For MRLs where the traditional assumption of reliable segmentation breaks down, results should be reported under both gold and automatic segmentation conditions, as well as on raw text, to provide realistic performance estimates.



\paragraph{Generative LLMs}

\begin{table}[t]
  \centering
  \small
  \begin{adjustbox}{max width=\columnwidth}
    \begin{tabular}{lcccc}
      \toprule
      \textbf{Model} & \textbf{Raw} & \textbf{Automatic} & \textbf{Gold} & \textbf{Gold} \\
       & \textbf{Text} & \textbf{Seg.} & \textbf{Seg.} & \textbf{Mentions} \\
      \midrule
      Dicta 2.0              &  1.0 &  1.5 &  0.3 & 13.8 \\
      GPT-4.1                & 15.1 & 17.2 & 17.7 & 44.8 \\
      GPT-4o                 & 14.2 & 15.3 & 14.5 & \textbf{45.4} \\
      o1                     & 13.4 & 16.1 & 17.9 & 37.9 \\
      o3                     & 15.7 & 18.8 & 19.4 & 42.1 \\
      Gemini 2.0-Flash       & 13.2 & 19.1 & 15.2 & 41.0 \\
      Gemini 2.0-Flash-Lite  & 12.1 & 14.8 & 15.2 & 38.4 \\
      Gemini 2.5-Pro         & \textbf{22.2} & \textbf{27.4} & \textbf{26.8} & 44.7 \\
      \midrule
      Best neural baseline   & --- & 47.9 & 50.4 & --- \\
      \bottomrule
    \end{tabular}
  \end{adjustbox}
  \caption{LLM performance (CoNLL F$_1$) under four regimes. Results averaged across 5 runs; closed-source LLMs exhibit $\sigma$=0.4--3.2 despite temperature 0, reflecting known non-determinism in production systems.}
      \label{tab:llm-results}
\end{table}

Table~\ref{tab:llm-results} evaluates  state-of-the-art LLMs on Hebrew CR using in-context learning under four evaluation conditions: raw text, automatic segmentation, gold segmentation, and gold mentions. 

Surprisingly, even when provided with perfect mention boundaries (the gold-mention condition), the best-performing LLM (GPT-4o at 45.4 F$_1$) falls 5.0 points behind the much smaller neural encoder-based baseline (50.4 F$_1$). This finding is particularly striking given the vast parameter difference—hundreds of billions compared with hundreds of millions. The underperformance persists across all evaluated models; prominent LLMs like o3, Gemini 2.5-Pro, and GPT-4.1 all fail to exceed 45 F$_1$. This pattern directly contradicts the English pattern where the same LLMs consistently outperform neural encoder-based systems (Table~\ref{tab:cross-language}). The inverse performance trend, from a +7.0 F$_1$ advantage in English to a -5.0 F$_1$ deficit in Hebrew, suggests that current LLMs struggle with Hebrew coreference under morphological complexity.

This underperformance highlights a compounded bottleneck: a combined failure in segmentation and clustering. Providing LLMs with gold mentions yields a massive performance jump (from 26.8 F$_1$ to 45.4 F$_1$), confirming that mention boundary recovery remains a significant hurdle. At the same time, the comparison between automatic and gold segmentation is mixed across LLMs, suggesting that better segmentation alone does not consistently resolve the problem. However, the fact that LLM performance plateaus at 45.4 F$_1$ even with perfect mention boundaries---still 5.0 points behind the much smaller neural encoders---indicates that a substantial bottleneck remains at the discourse-level clustering stage.

Our detailed error analysis further supports this interpretation. While gold segmentation inflates pronoun share, their resolution is often easier via proximity and agreement features and therefore yields only modest gains. However, neural encoders produce five times more correct clusters than LLMs (9.2/8.9 vs.\ 1.6–1.9 per document) and miss seven fewer gold clusters (9.7–10.0 vs.\ 17.0–17.3). Thus, even when both architectures receive segmented input, LLMs continue to struggle with discourse-level clustering. See further details in Appendix \ref{sec:error-analysis}.


\begin{table}[t]
  \centering
  \begin{adjustbox}{max width=\columnwidth}
  \begin{tabular}{lccc}
    \toprule
    \textbf{Language} & \textbf{Neural-Encoder} & \textbf{LLM} & \textbf{$\Delta$} \\
     & \textbf{(Gold Seg.)} & \textbf{(Gold Mentions)} & \textbf{(LLM$-$NE)} \\
    \midrule
    English & 81.4 & 88.4 & \textbf{+7.0} \\
    Hebrew  & 50.4 & 45.4 & \textbf{$-$5.0} \\
    \bottomrule
  \end{tabular}
  \end{adjustbox}
  \caption{Cross-linguistic performance inversion. English results from \citet{le2023largelanguagemodelsrobust} using the same LingMess architecture. The 12-point swing between languages reveals fundamental limitations of current LLMs on morphologically rich languages.}
  \label{tab:cross-language}
\end{table}

\section{Conclusion}
\label{sec:conclusion}
This paper introduces {\em KibutzR}, the first coreference corpus for Modern Hebrew, and uses it to re-examine modeling and evaluation practices of coreference resolution under morphological complexity. By evaluating both supervised encoders and frontier LLMs, across scenarios with raw text, automatic segmentation, gold segmentation, and gold mentions, we can  isolate and characterize the segmentation and clustering challenges.

We show that, first, state-of-the-art segmentation reduces performance of neural encoders  by 1.4--2.6 F$_1$ points relative to gold. Next, contemporary LLM decoders underperform neural encoders by 5 points in Hebrew, even with gold mentions, reversing the English pattern where  decoders dominate. Finally, we show that LLMs face a combined bottleneck in boundary recovery and clustering: while gold mentions yield dramatic improvements ($\sim$20--30 F$_1$), the comparison between automatic and gold segmentation is mixed across LLMs, suggesting that segmentation alone does not explain the gap, and that substantial difficulty remains at the clustering stage.

These findings reveal that current CR modeling and prompting practices struggle with Hebrew under the dual challenge of segmentation and coreference.
We release KibutzR, its guidelines, its annotation UI, prompts, and evaluation code, to enable research toward architectures that better handle morphological segmentation and coreference resolution as interconnected components, in Hebrew and other MRLs.

\section*{Limitations}

Our corpus is limited to newswire text; broader genres remain a subject of future work. The requirement for UD-annotated text during annotation restricted us to news. The use of automatic parses may mitigate this, but risk  increasing error propagation sources. Recently released Hebrew UD parsers and analyzers  enable expansion to other domains.

The corpus consists of publicly available news articles. 
As newswire may encode topical and gender biases, results may not generalize beyond news domains, or across news domains in vastly different (temporally spread) eras.

LLM results reflect zero-shot prompting for gold mentions (following prior work) and few-shot prompting for end-to-end evaluation; alternative instruction curricula may improve performance, but do not eliminate the need to model segmentation uncertainty. Finally, while our automatic segmenter is strong, improved segmentation tools could narrow (even if not entirely erase) the observed gaps. 

Finally, while all of our code, guidelines, modeling, and experimental design is done in a language-agnostic manner, it is executed and evaluated only on Modern Hebrew texts. Parallel stream of research on additional MRLs are needed to strengthen the cross-lingual manifestation of this challenge. We hope that are code, guidelines, and actual tools (UI, evaluation setups) will greatly facilitate and expand the development of such resources and analyses for multiple languages.

\section*{Acknowledgments}
We thank Omer Goldman and Arie Cattan for their insightful comments, and three anonymous reviewers for their valuable feedback. This research was supported  by a grant from the Israeli Science Foundation (ISF grant no.\ 670/23) as well as a grant from the Israeli Innovation Authority (KAMIN), for which we are grateful. The  computing resources for the project were kindly funded by a VATAT grant from the  Planning and Budgeting Committee of the Council for Higher Education in Israel.
\bibliography{acl2025}

\input{appendix_clean}

\section{Error Analysis}
\label{sec:error-analysis}

We analyze where coreference models fail in Hebrew by comparing error patterns across segmentation conditions and model architectures. We examine (1) how mention type distributions --- particularly pronoun ratios --- shift with segmentation methods, and (2) clustering success rates. We categorize model outputs into three outcomes: \emph{Correct} clusters (predicted clusters that match gold clusters), \emph{Extra} clusters (spurious predictions), and \emph{Missed} clusters (unresolved gold clusters). To determine matches, we compute weighted overlap between predicted and gold clusters, down-weighting pronouns (0.2) versus content mentions (1.0).\footnote{Multiple entities can share the same pronoun form (e.g., multiple male entities all referred to as "he"), making pronoun overlap weak evidence for cluster matching.}

\paragraph{Experimental Setup.} 
We compare five configurations: \textbf{Neural–Gold} and \textbf{Neural–SOTA} (our best-performing neural encoder based system on gold/automatic segmentation), \textbf{LLM–Gold}, \textbf{LLM–SOTA}, and \textbf{LLM–Raw} (Gemini~2.5~Pro on gold/automatic/raw text). All analyses use the dev split.

\begin{figure}[t]
  \centering
  \includegraphics[width=.98\columnwidth]{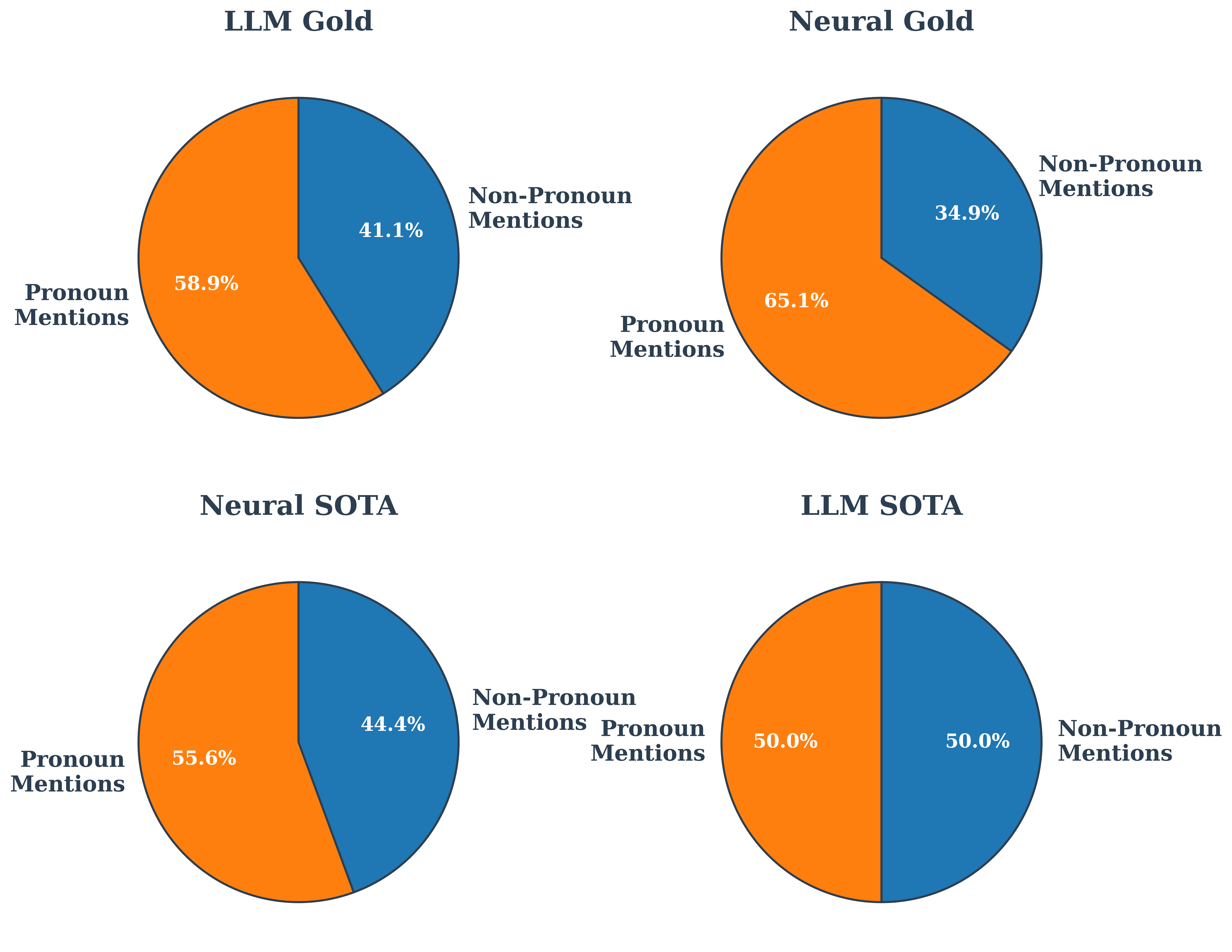}
  \caption{\textbf{Pronoun share rises under gold segmentation.}
  \textbf{LLM}: 50.0\% (\textit{LLM–SOTA}) $\rightarrow$ 58.9\% (\textit{LLM–Gold}) (\,+8.9pp).
  \textbf{Neural}: 55.6\% (\textit{Neural–SOTA}) $\rightarrow$ 65.1\% (\textit{Neural–Gold}) (\,+9.5pp).}
  \label{fig:mention-pies-4}
\end{figure}

\paragraph{Gold segmentation increases pronoun recovery.}
By examining which mentions each segmentation method discovers, we observe that switching from automatic to gold segmentation increases pronoun mention share by about 9 percentage points in both architectures (Figure~\ref{fig:mention-pies-4}). This increase occurs because gold segmentation more reliably recovers grammatical morphemes, including bound pronouns such as possessive suffixes and object clitics, that automatic segmenters may miss or attach incorrectly. Many of these recovered pronouns are easier to resolve using local agreement and proximity cues, so their increased availability likely contributes to the gains under gold segmentation. At the same time, this pattern suggests that improved segmentation does not primarily recover the hardest content mentions.

\begin{figure}[t]
  \centering
  \includegraphics[width=.98\columnwidth]{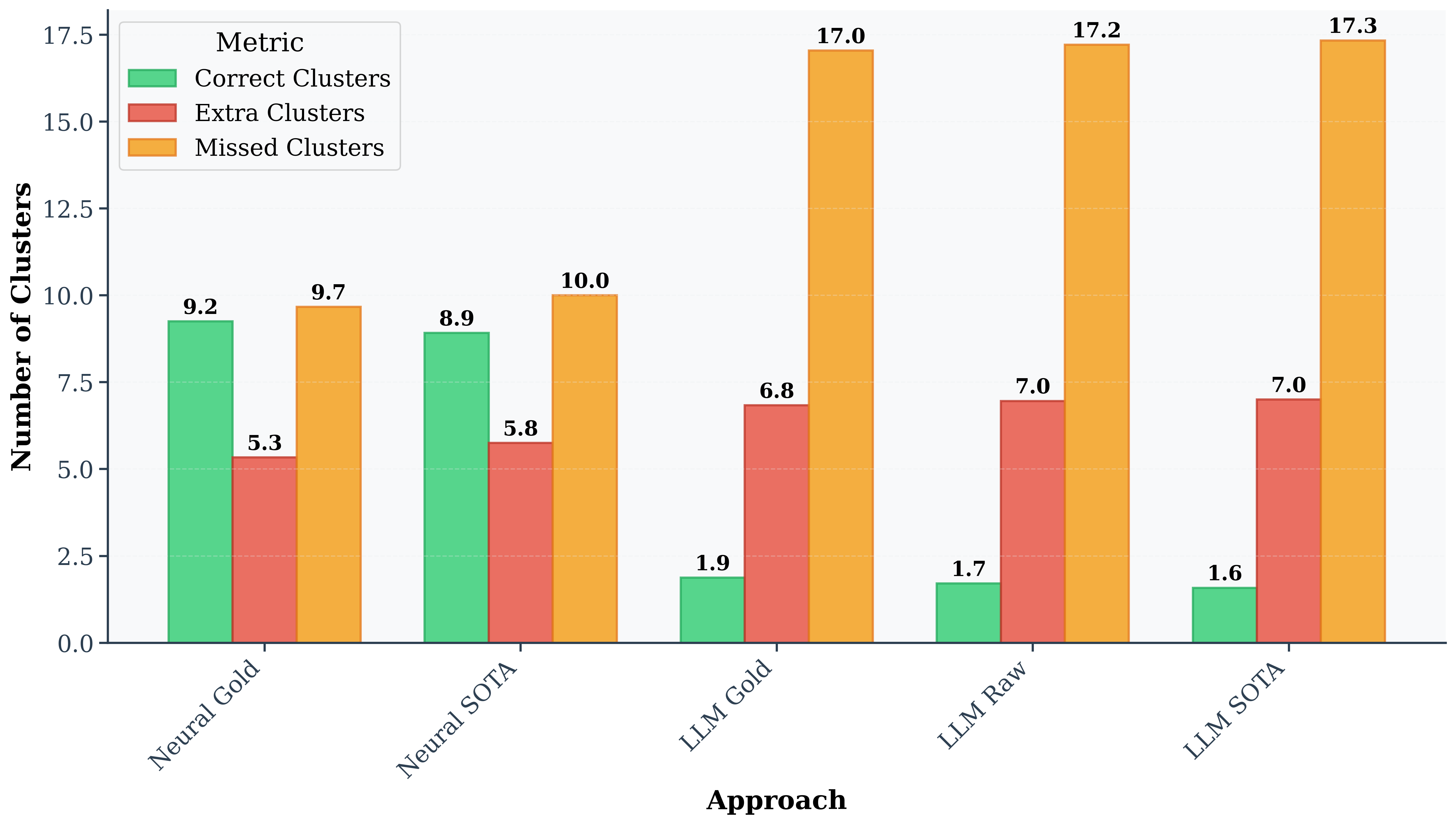}
  \caption{\textbf{Cluster outcomes by approach (per-document means).}
  Neural models produce ~5× more correct clusters than LLMs regardless of segmentation quality.}
  \label{fig:cluster-bars}
\end{figure}

\paragraph{Improved segmentation alone does not close the LLM gap.}
Figure~\ref{fig:cluster-bars} reveals a striking architecture gap. Neural encoders produce about five times more correct clusters than LLMs (9.2/8.9 vs.\ 1.6--1.9 per document) and miss roughly seven fewer gold clusters (9.7--10.0 vs.\ 17.0--17.3). Importantly, this gap persists even when both architectures receive segmented input (Neural--SOTA vs.\ LLM--SOTA), indicating that better boundary information alone is insufficient to close the gap.

The three LLM conditions (Raw/SOTA/Gold) reinforce this interpretation. Segmented input improves performance relative to raw text, but correct clusters remain sparse across all LLM settings. Taken together, these analyses suggest that segmentation is one important source of difficulty, while a substantial residual bottleneck remains at the discourse-level clustering stage.

\end{document}

%% file: appendix_clean.tex
\clearpage
\appendix

\section{Mention-Detection Implementation}
\label{app:mention-detection}

\subsection{Inputs and Dependencies}
The system requires Universal Dependencies parses of Hebrew text, including tokenization, POS tags, and dependency labels. Root candidates comprise \texttt{NOUN}, \texttt{PROPN}, \texttt{PRON}, \texttt{NUM}, and verbs when occurring in \textit{smixut} constructions. Span expansion utilizes the following permitted dependency labels: \texttt{appos}, \texttt{compound:smixut}, \texttt{nmod:poss}, and \texttt{conj}.

\subsection{Span Construction Rules (Deterministic)}
Spans expand bidirectionally from a root node along the permitted dependency edges, optionally including \texttt{amod} and \texttt{det} modifiers according to Hebrew-specific rules. The system handles Hebrew-specific phenomena including pronominal clitics (possessive, object, and prepositional), nested \textit{smixut} constructions, determiners, numerals, and conjunctions. When overlaps occur, the system resolves them by selecting the maximal span, while compatible overlaps are merged. The traversal process is fully deterministic, employing fixed token order and single-threaded graph walks to ensure reproducibility.

\subsection{Mention-Detection Pseudocode}
\noindent\textit{Preconditions.} UD-parsed Hebrew sentences; root candidates:
\texttt{NOUN/PROPN/PRON/NUM} and verbs-in-\textit{smixut}; permitted edges:
\texttt{appos}, \texttt{compound:smixut}, \texttt{nmod:poss}, \texttt{conj}.
Deterministic traversal; overlaps resolved by maximal span.

{%
\setlength{\textfloatsep}{0.6\baselineskip}%
\begin{algorithm}[ht]
\small
\caption{High-level Mention Detection for Hebrew}
\label{alg:hebrew-md}
\begin{algorithmic}[1]
\FOR{each sentence in a UD-parsed corpus}
  \FOR{each token $t$ in the sentence}
    \IF{$t$ is a root candidate (NOUN/PROPN/PRON/NUM/verb-in-\textit{smixut})}
      \STATE Initialize span $S \gets [t]$
      \STATE Expand $S$ left/right via permitted dependency relations
      \STATE Apply Hebrew-specific rules (clitics, determiners, numerals, conjunctions)
      \STATE Add $S$ to mention candidates
    \ENDIF
  \ENDFOR
\ENDFOR
\STATE Filter/merge overlaps by maximal span
\STATE Output all candidate mentions
\end{algorithmic}
\end{algorithm}
}

\noindent The complete Mention Detection system will be released upon publication.

\section{Prompt Templates}
\label{app:prompts}

\paragraph{Gold-Mention Clustering (\texttt{doc\_template}).}
\noindent\fbox{%
\parbox{\columnwidth}{\footnotesize
\textbf{Task:} Annotate all entity mentions in the following text with coreference clusters.\\
Use Markdown tags to indicate clusters, e.g. \texttt{[mention](\#cluster\_name)}.\\
Do not add extra information or clusters beyond those marked.

\textbf{Input}: [Tom](\#) and [Mary](\#) go to [the park](\#). [It](\#) was full of trees.\\
\textbf{Output}: [Tom](\#cluster\_0) and [Mary](\#cluster\_1) go to [the park](\#cluster\_2). [It](\#cluster\_2) was full of trees.
}}

\paragraph{End-to-End Coreference (\texttt{e2e\_template}).}
\begin{tcolorbox}[enhanced,breakable,
    colback=white,colframe=black,boxrule=0.3pt,arc=2mm,
    left=2mm,right=2mm,top=1mm,bottom=1mm,
    fontupper=\footnotesize,width=\columnwidth,sharp corners=south]
Cluster all entity mentions in the following Hebrew text to coreference clusters.
Use Markdown tags to indicate the coreference group in the output, with the
format \texttt{[mention](\#)}, e.g.\\
\texttt{\texthebrew{כש[הם](\#) הלכו ל[בית של [אנחנו](\#)](\#)}}

\medskip
\texttt{• First, tokenize words to expose clitics}\\
\texttt{• Second, mark the mentions}\\
\texttt{• Finally, cluster all mentions together}

\texttt{Tokenization example}\\
\texttt{\# Input word: \texthebrew{ביתנו}} \\
\texttt{\# Output: \texthebrew{בית\_של\_אנחנו}} \\
\texttt{\# Input word: \texthebrew{העסקתם}} \\
\texttt{\# Output: \texthebrew{העסקה\_של\_הם}}

\medskip
\texttt{Mention-marking example}\\
\texttt{\# Input tokenized: \texthebrew{הוא איבד את הכרה\_של\_הוא}}\\
\texttt{\# Output: \texthebrew{[הוא](\#) איבד את [הכרה\_של\_[הוא](\#)](\#)}}

\medskip
\texttt{Clustering example}\\
\texttt{\# Input: \texthebrew{[הוא](\#) איבד את [הכרה\_של\_[הוא](\#)](\#)}}\\
\texttt{\# Output: \texthebrew{[הוא](\#אגד\_1) איבד את [הכרה\_של\_[הוא](\#אגד\_1)](\#אגד\_2)}}

\medskip
\textbf{Examples}\\
\texttt{\# Input:}\\
\texthebrew{לידס עלתה למקום החמישי אחרי שניצחה אתמול בחוץ במשחק השלמה את מנצסטר סיטי 3 2. השערים ללידס: לי צפמאן (14), קארל שאט (42), גורדון סטראקאן (62). לסיטי: אשלי וורד (49 מ-11 מ), דייויד ווייט (65).}\\
\texttt{\# Output:}\\
\texthebrew{[לידס](\#אגד\_0) עלתה ל ה\_ מקום ה חמישי אחרי ש ניצחה אתמול ב ה\_ חוץ ב משחק השלמה את [מנצסטר סיטי](\#אגד\_1) 3 2. [[ה שערים ל[לידס](\#אגד\_0)](\#):[לי צפמאן (14), קארל שאט (42), גורדון סטראקאן (62)]](\#). ל[סיטי](\#אגד\_1): אשלי וורד (49 מ - 11 מ), דייויד ווייט (65).}

\medskip
\texttt{\# Input:}\\
\texthebrew{הרבה החמצות ממצבים נוחים של יבנה, בגלל משחק הגנתי של טבריה שהזמינה התקפות. בין חלוצי יבנה, שהרבו להחמיץ, ניצל אנריקה ורון הזדמנות אחת בלבד, כדי להעניק לקבוצתו פרס של 3 נקודות בעד נצחון שהיתה ראויה לו. שפט אריה וולף, 000{,}1 צופים, ביבנה.}\\
\texttt{\# Output:}\\
\texthebrew{הרבה החמצות מ מצבים נוחים של [יבנה](\#אגד\_0), בגלל משחק הגנתי של טבריה ש הזמינה התקפות. בין חלוצי [יבנה](\#אגד\_0), ש הרבו להחמיץ, ניצל [אנריקה ורון](\#אגד\_1) הזדמנות אחת בלבד, כדי להעניק ל[קבוצה\_של\_[הוא](\#אגד\_1)](\#אגד\_0) פרס של 3 נקודות בעד [נצחון ש היתה ראויה ל[הוא](\#אגד\_0)](\#אגד\_1). שפט אריה וולף, 000{,}1 צופים, ב יבנה.}

\medskip
\texttt{Nested mentions are allowed; mark every nested span and noun-phrase candidate.}\\
\texttt{Keep the text exactly as it was, except for Markdown.}\\
\texttt{Do not output singletons in the final cluster document.}
\end{tcolorbox}

\section{LLM Inference Pipeline}

\subsection{Deterministic Settings}
\begin{itemize}[noitemsep,leftmargin=*]
  \item Temperature $=0.0$; default top-$p$; max output tokens $=\min(\text{context\_limit}-\text{input}, 4096)$.
  \item Up to 3 retries on format-validation failure; fixed random seeds where applicable.
\end{itemize}

\begin{algorithm}[ht] 
\small
\caption{Inference and Output Validation}
\begin{algorithmic}[1]
\FOR{each document $d$}
 \STATE $p \leftarrow$ \texttt{render\_template}$(d)$
 \STATE $r \leftarrow$ \texttt{model}$(p,$ temperature $=0)$
 \STATE Validate bracket balance, sentence alignment, cluster-ID schema
 \IF{validation fails}
   \STATE apply non-semantic formatting fix
   \STATE retry (max 3)
 \ENDIF
 \STATE Parse clusters $\rightarrow\ C_d$
\ENDFOR
\STATE \textbf{return} $\bigcup_d C_d$
\end{algorithmic}
\end{algorithm}

\subsection{Models and Artifacts}
\begin{itemize}[noitemsep,leftmargin=*]
  \item Models used (replication list): GPT-4o, GPT-4o-mini, GPT-4.1, o1, o3, Gemini 2.5 Pro, Gemini 2.0 Flash, Gemini 2.0 Flash Lite, DictaLM 2.0-Instruct.
  \item The experimental framework is built on OpenAI API v0.28.1 and vLLM for local model serving.
\end{itemize}

\section{Neural Baseline Configurations}
\label{app:neural-hyperparams}

\subsection{Training Regime}
\begin{itemize}[noitemsep,leftmargin=*]
  \item Optimizer: AdamW; dropout: 0.3; weight decay: 0.01; epochs: 150.
  \item Seeds for averaging: \{42, 123, 2021, 27182, 31415\}.
  \item Max segment length: 512; max span width: 30 (LingMess-He) / 64 (wl-coref); rough scoring $k=50$ (wl-coref).
\end{itemize}

\begin{table}[H]
\centering
\small
\setlength{\tabcolsep}{4pt}
\begin{adjustbox}{max width=\linewidth}
\begin{tabular}{lcc}
\toprule
\textbf{Hyperparameter} & \textbf{LingMess-He} & \textbf{wl-coref} \\
\midrule
LR (encoder)               & $1\times 10^{-5}$ & $1\times 10^{-5}$ \\
LR (task)                  & $3\times 10^{-4}$ & $3\times 10^{-4}$ \\
Dropout                    & 0.3               & 0.3               \\
Max span width             & 30                & 64                \\
Max segment length         & 512               & 512               \\
Rough scoring $k$          & --                & 50                \\
Hidden size (FFNN/Scorer)  & 2048              & 1024              \\
Weight decay               & 0.01              & 0.01              \\
\bottomrule
\end{tabular}
\end{adjustbox}
\caption{Key hyperparameters for neural baselines (replication).}
\label{tab:neural-hparams}
\end{table}

\section{Annotation Guidelines (Operational)}
\subsection{Mention Types}
Named entities; nominals; pronominals (including clitics); zero pronouns when recoverable from morphology.

\subsection{Hebrew-Specific Handling}
\begin{itemize}[noitemsep,leftmargin=*]
  \item Pronominal clitics: possessive on nouns (\texthebrew{ספרו} $\rightarrow$ \texthebrew{[ספר\_של\_[הוא]]}), object on verbs (\texthebrew{ראיתיו} $\rightarrow$ \texthebrew{ראיתי\_[אותו]}), prepositional clitics (\texthebrew{אליהם} $\rightarrow$ \texthebrew{אל\_[הם]}).
  \item Construct state (\textit{smixut}): allow nested constructs; annotate construct and internal heads when referential.
\end{itemize}

\subsection{Exclusions}
We exclude pleonastic or dummy uses, idioms with non-referential components (e.g., \texthebrew{יצא\ מכליו}), and negated non-existent entities (e.g., \texthebrew{אין\ לי\ מכונית}).

\subsection{Inter-Annotator Agreement Protocol}
The annotation process began with initial calibration on 10 documents followed by group discussion. We then conducted three annotation batches with continuous agreement monitoring and targeted rule clarifications. Final expert adjudication resolved disagreements, with IAA computed on pre-adjudication annotations.

\section{Scoring Definitions}
\subsection{Weighted Similarity}
For brevity let $C_p=C_{\text{pred}}$ and $C_g=C_{\text{gold}}$.
\begin{equation}
\mathrm{Sim}(C_p, C_g) =
\frac{\sum_{m \in C_p \cap C_g} w(m)}{\sum_{m \in C_p \cup C_g} w(m)}.
\end{equation}
\noindent We assign weights based on mention type: pronouns (including clitics) receive $w(m)=0.2$, while content mentions (nouns and named entities) receive $w(m)=1.0$.

\subsection{Outcome Categories}
\begin{itemize}[noitemsep,leftmargin=*]
  \item \textbf{Correct}: similarity $\ge 0.5$
  \item \textbf{Extra}: predicted cluster without a sufficient gold match
  \item \textbf{Missed}: gold cluster without a sufficient predicted match
\end{itemize}

\section{Environment and Artifact Bundle}
Our experiments support single-GPU training with CPU-only inference capability. The artifact bundle includes all tool and model versions, encoder checkpoints, random seeds, run scripts, prompt JSON files, validator regexes, and parsing scripts. The bundle will be distributed upon publication.